\documentclass[a4paper, conference, final]{IEEEtran}

\makeatletter
\def\bstctlcite{\@ifnextchar[{\@bstctlcite}{\@bstctlcite[@auxout]}}
\def\@bstctlcite[#1]#2{\@bsphack
  \@for\@citeb:=#2\do{%
    \edef\@citeb{\expandafter\@firstofone\@citeb}%
    \if@filesw\immediate\write\csname #1\endcsname{\string\citation{\@citeb}}\fi}%
  \@esphack}
\makeatother

\usepackage[vlined, ruled, linesnumbered, commentsnumbered]{algorithm2e}
\usepackage{amsmath}
\usepackage{cite}
\usepackage{color}
\usepackage{xcolor}
\definecolor{chred}{rgb}{0.8,0,0}
\definecolor{chgray}{rgb}{0.5,0.5,0.5}
\usepackage{graphicx}
\usepackage{caption}
\usepackage{subcaption}
\usepackage{dblfloatfix}
\usepackage{eqlist}
\usepackage{txfonts}
\usepackage{url}
\usepackage{footmisc}
\usepackage{booktabs}
\usepackage{balance}

\usepackage{multirow}
\usepackage{threeparttable}
\usepackage{geometry}
\begin{document}
\newgeometry{top=25.4mm, bottom=19.1mm, left=19.1mm, right = 19.1mm} 
\IEEEoverridecommandlockouts

\title{\LARGE \bf
A Double Jaw Hand Designed for Multi-object Assembly}

\author{Joshua C. Triyonoputro$^{1}$, Weiwei Wan$^{1,2,*}$, Kensuke
Harada$^{1,2}$
\thanks{$^{1}${Graduate School of Engineering Science, Osaka University, Japan.}
$^{2}${National Inst. of AIST, Japan.} {*Correspondent author: Weiwei Wan}
{\tt\small wan@hlab.sys.es.osaka-u.ac.jp}} }

\maketitle
\thispagestyle{empty}
\pagestyle{empty}

\begin{abstract}
This paper presents a double jaw hand for industrial assembly. The
hand comprises two orthogonal parallel grippers with different mechanisms.
The inner gripper is made of a crank-slider mechanism which is compact and able
to firmly hold objects like shafts. The outer gripper is made of a parallelogram
that has large stroke to hold big objects like pulleys.
The two grippers are connected by a prismatic joint along the hand's
approaching vector. The hand is able to hold two objects and perform in-hand
manipulation like pull-in (insertion) and push-out (ejection).
This paper presents the detailed design and implementation of the hand, and
demonstrates the advantages by performing experiments on two sets of
peg-in-multi-hole assembly tasks as parts of the World Robot Challenge (WRC)
2018\footnote{http://worldrobotsummit.org/en/programs/challenge/} using a
bimanual robot.
\end{abstract}

\section{Introduction} \label{introduction}

\begin{figure}[!t]
        \centering
        \begin{subfigure}{\columnwidth}
                \centering
                \includegraphics[width=\columnwidth]{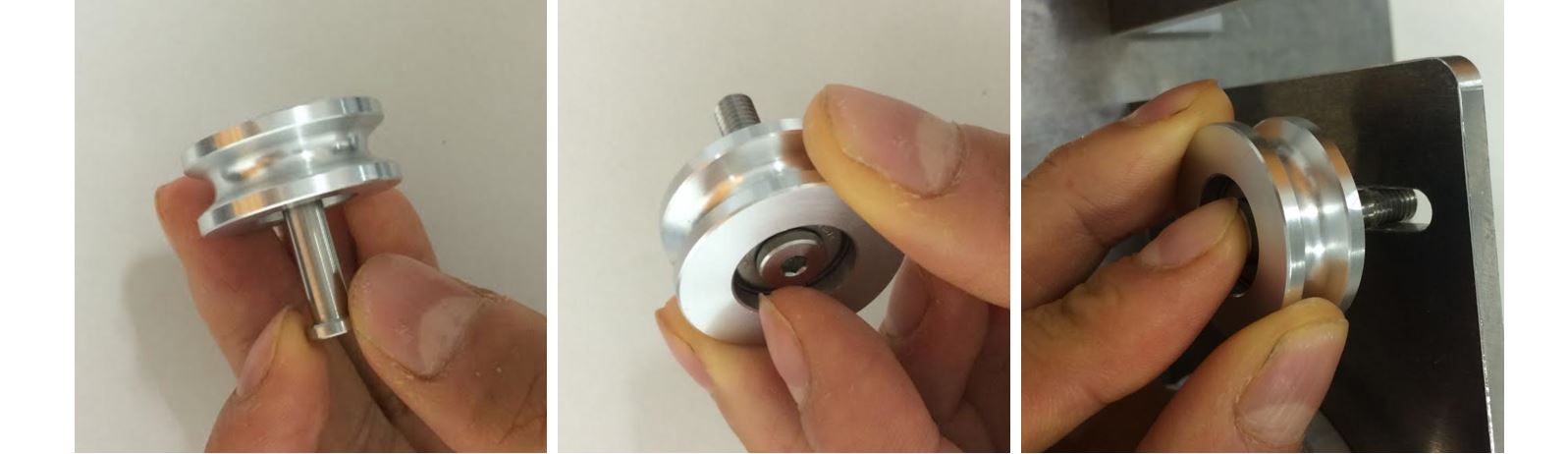}
                \caption{A human hand holding and manipulating two objects with
                one hand in product assembly.}
                \label{fig:intr_a}
        \vspace{0.05in}
        \end{subfigure}
        ~
        \begin{subfigure}{\columnwidth}
                \centering
                \includegraphics[width=\columnwidth]{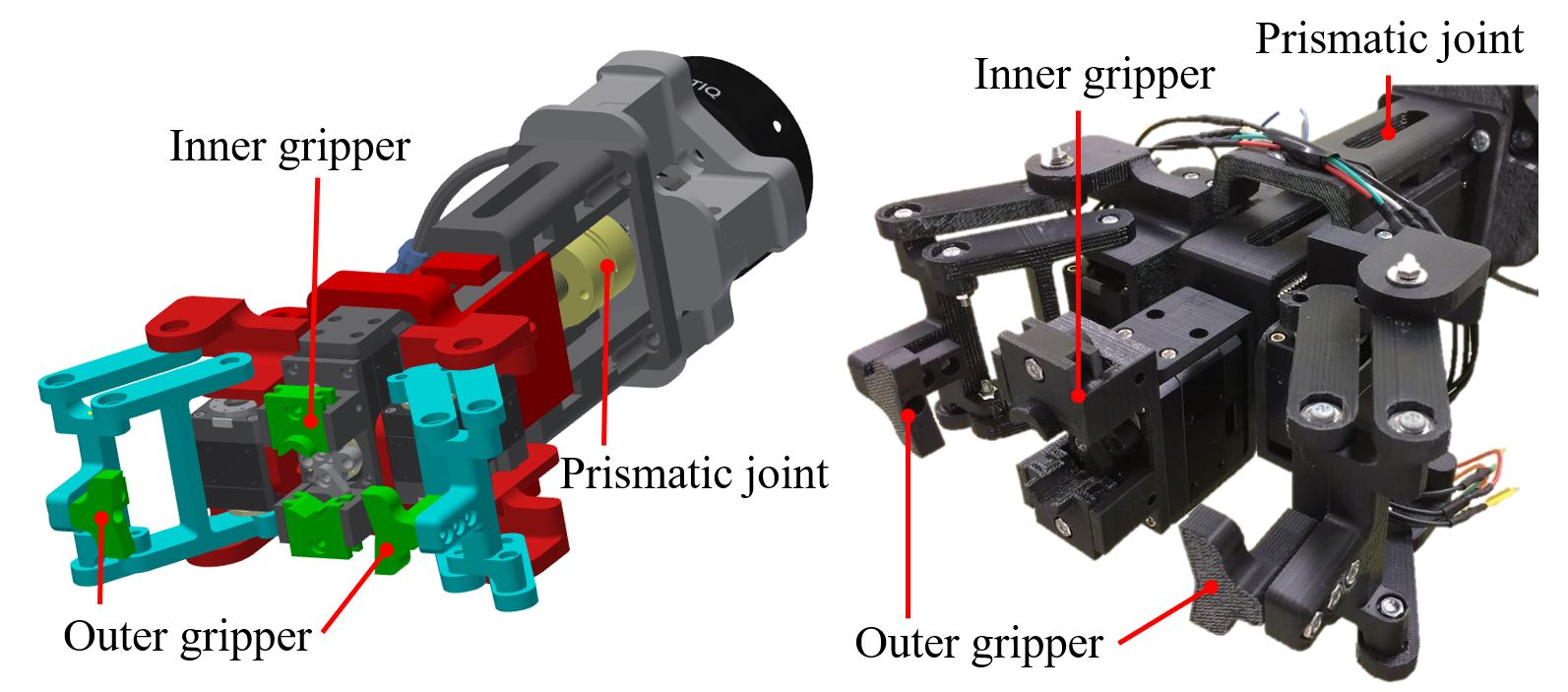}
                \caption{CAD models of the double jaw hand and an
                implementation.}
                \label{fig:intr_b}
        \end{subfigure}
        \caption{Motivations of the work and the developed
        hand.}
        \label{teaser}
\end{figure}

One major challenge of next-generation manufacturing is autonomous
assembly. Many studies have been devoted to related problems
in the past decades \cite{hormann1991development}, \cite{dogar2015multi},
\cite{wan2017teaching}. Most of them used parallel grippers to
plan grasps and assembly sequences. Although they made impressive progress, 
an inherent problem remains -- holding and manipulating multiple objects
simultaneously. Fig.\ref{teaser}(a) shows an example where a human hand holds
and manipulates two objects (a peg and a pulley) together. The task is widely
seen in real-world product assembly and difficult to be performed using simple
parallel grippers.

There are several possible hand design solutions to the remaining problem. For
example, some designs used two or more grippers in one robotic hand
\cite{monkman2007robot}. The grippers can be fixed to tuning turrets
\cite{Mason01}, or they can have one or more Degree of Freedoms (DoFs) relative
to each other \cite{zeng2017robotic}, \cite{atakuru2018robotic}.
Some other designs used fully actuated \cite{xu2016design} or underactutated
anthropomorphic hands \cite{townsend2000barretthand}, \cite{massa2002design},
\cite{deimel2016novel}.
Specifically, Zeng et al. \cite{zeng2017robotic} developed a gripper with a
retractable mechanism to allow switching between a parallel gripper and a
suction gripper. Cannella et al.
\cite{cannella2013design} and Chen et al. \cite{chen2014hand} developed
industrial grippers with twisting ability for high-speed assembly. Ma et al.
\cite{ma2016m} developed a two-finger gripper using an underactuated
human-like finger and a passive thumb. It is capable of performing in-hand
manipulation like pull-in and push-out. Odhner et al.
\cite{odhner2014compliant} developed a 3-finger underactuated hand using a
minimalistic design optimized for a set of tasks. Yamaguchi et al.
\cite{yamaguchi2013development} added dexterity to underactuated
fingers by adding suction cups on each fingertip. Kakogawa et al.
\cite{kakogawa2016underactuated} developed an under-actuated three-finger
gripper with pull-in ability. Chavan-Dafle et al.
\cite{dafle2014extrinsic} used a 3-finger one-parameter gripper to extrinsically
manipulate objects. The finger shape of the gripper was optimized for holding
spherical objects \cite{rodriguez2013effector}. These designs suggested general
solutions to tackle the problem of holding and manipulating multiple objects
simultaneously, but they do not fully address the undergoing details. They
either only solve the problem partially or have special mechanisms that decrease
the robustness of robotic systems.

The inherent problem and the drawbacks of the available hand designs inspire us
to develop a more dexterous but still simple robotic hand for autonomous assembly.
The requirements of the new hand are as follows.
\begin{enumerate}
  \item Simple mechanisms and small number of actuators
  \item Capable of holding two objects
  \item Capable of in-hand pull-in and push-out
  \item Capable of aligning objects along a common axis
  \item Large stroke and large holding force
\end{enumerate}

In response to the requirements, we propose a double jaw hand that satisfies
these requirements.
The CAD models and an implementation are shown in Fig.\ref{teaser}(b).
The hand is made of two orthogonal parallel grippers with different
mechanisms. The inner gripper is made of a crank-slider mechanism which is
compact and able to firmly hold objects like shafts. The outer gripper is made
of a parallelogram that has large stroke to hold big objects like pulleys. The
two grippers are connected by a prismatic joint along the hand's
approaching vector. The hand is able to hold two objects and perform in-hand
manipulation like pull-in (insertion) and push-out (ejection).

The design is verified with real-world executions of peg-in-multi-hole
assembly tasks using objects from the World Robotic Challenge (WRC) 2018
assembly challenge. Experiments are also performed to show that the hand is
able to align a peg and a pulley, and assemble the peg into the
pulley without external help.

The paper is organized as follows. Section \ref{problem_analysis} explains in
details the motivations and the requirements for designing the hand. Section
\ref{gripper_design} presents the details of the gripping mechanisms and implementations.
Section \ref{experiments} demonstrates the efficacy of the developed hand
using two tasks from the WRC2018 assembly challenge. Section \ref{conclusion}
draws conclusions and presents future work.

\section{Motivations} \label{problem_analysis}

The proposed design is mainly motivated by the tasks in
WRC2018 assembly challenge.

\begin{figure}[!htbp]
\includegraphics[width=\columnwidth]{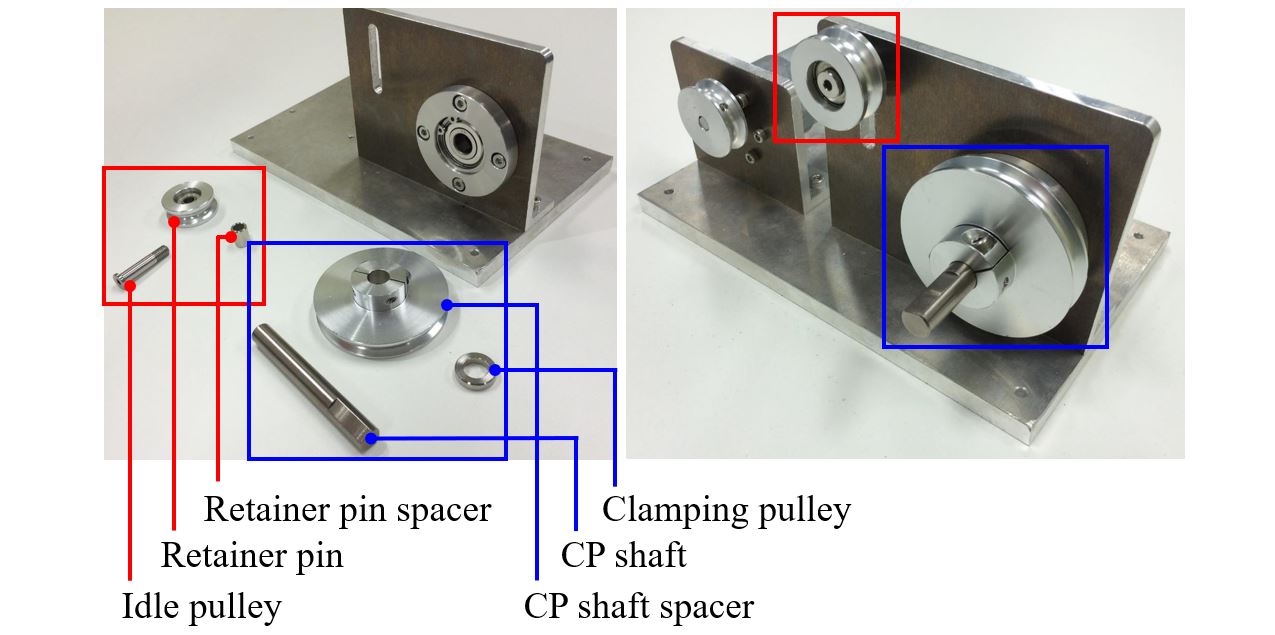}
\caption{Objects used in the peg-in-multi-hole assembly tasks of the WRC
2018 assembly challenge. The objects in red and blue boxes will be assembled
to a base respectively.}
\label{fig:objects}
\end{figure}

WRC2018 assembly challenge requires using robots to autonomously
assemble a Belt Drive Unit shown in Fig.\ref{fig:objects}(b). This paper
focuses on tackling one aspect of the whole task -- peg-in-multi-hole
assembly. Peg-in-multi-hole assembly is an extension to peg-in-hole
assembly. The goal of peg-in-hole assembly is to insert a peg into a hole
\cite{inoue1971computer}, \cite{mason1981compliance}, \cite{zheng2017peg},
while the goal of a peg-in-multi-hole assembly is to continuously insert
peg-and-hole complex into other holes. In the WRC2018 assembly challenge, robots
have to perform several sets of peg-in-multi-hole assembly subtasks. Two of
them, namely the clamping pulley set and the idle pulley set, are shown in
Fig.\ref{fig:objects}(a). The clamping pulley set comprises inserting a pulley
shaft into a clamping pulley, inserting the shaft-and-pulley complex
into the hole of a pulley shaft spacer, and inserting the shaft-pulley-spacer
complex into the hole of a housing.
The idle pulley set comprises inserting a retainer pin into an idle pulley,
inserting the pin-and-pulley complex into a retainer pin spacer, and inserting
the pin-pulley-spacer complex into a slot.

\begin{figure}[!htbp]
\includegraphics[width=\columnwidth]{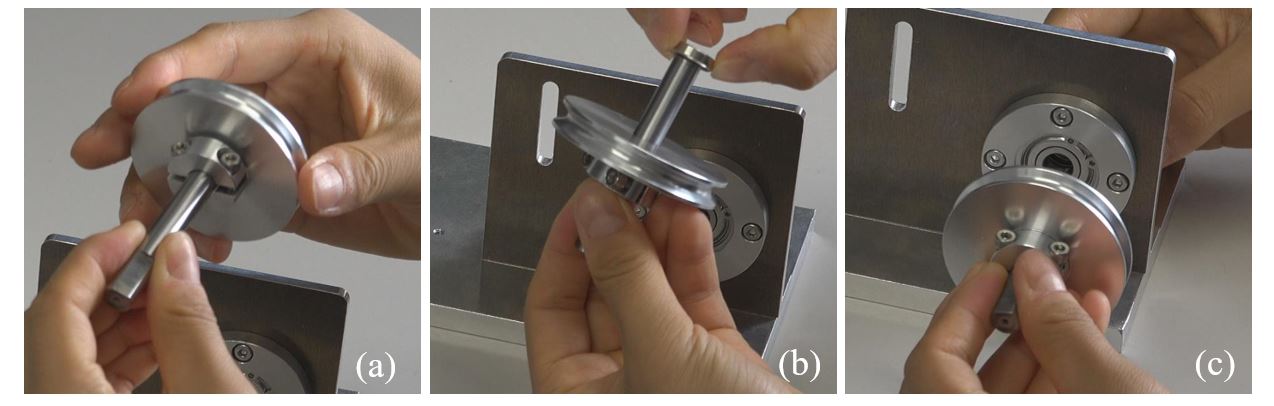}
\caption{Manually assembling the clamping pulley set. (a) Peg-in-hole: Insert a
shaft into the hole of a pulley. (2) Complex-in-hole: Inserting the
shaft-and-pulley complex into the hole of a spacer.
(3) Complex-in-hold: Insert the shaft-pulley-spacer complex into the hole of a
bearing.}
\label{fig:steps1}
\end{figure}

\begin{figure}[!htbp]
\includegraphics[width=1.0\columnwidth]{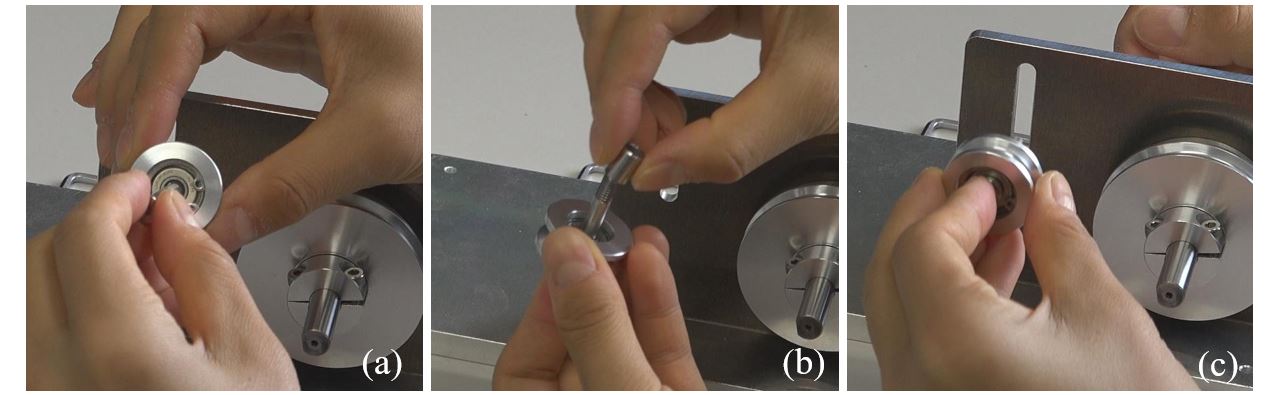}
\caption{Manually assembling the idle pulley set. (a) Peg-in-hole. (b)
Complex-in-hole: Insert the pin-and-pulley complex into the hole of a spacer.
(3) Complex-in-hole: Insert the pin-pulley-spacer complex into a slot on the
base.}
\label{fig:steps2}
\end{figure}

Essentially, the two sets of peg-in-multi-hole assembly share the same workflow,
which involves (a) inserting a peg to the hole of a pulley, (b) inserting
the peg-and-pulley complex to the hole of a spacer, (c) inserting the
peg-pulley-spacer complex into the hole on a metal base. Illustrations of human
assembling the two sets using the workflow are shown in Fig. \ref{fig:steps1}
and Fig. \ref{fig:steps2} respectively.

In order for a robotic hand to perform similar tasks, it must meet the
following requirements. First, the hand must be able to hold two objects to
avoid slipping while performing assembly. In Fig.
\ref{fig:steps1}(b, c), the inserted shaft must be held together with the pulley
and spacer after being inserted. Or else, the pulley and spacer may slip out of
the complex when the robot rotates its arm to move from one key pose to another,
leading failure in assembly. Another example is inserting the shaft into the
pulley. It is possible to finish the task using two arms with simple grippers:
One arm holds the clamping pulley, and the other holds the pulley
shaft. The pulley shaft is inserted into the pulley using impedance
control. However, the two arms in this configuration cannot directly continue to
the next peg-in-hole step after finishing the insertion. It has to perform some
regrasps and handovers: First, one arm releases the pulley shaft while
the clamping pulley is held by the other arm. Then, the shaft-and-pulley complex
is handed over to the free arm. Third, the two arms perform another handover to
let the previous arm hold one end of the shaft. The process is time consuming
and vulnerable. Errors and noises accumulate during the multiple times of
regrasp and handover, leading to failure in assembly.
Developing a hand that can hold two objects
could avoid these problems. For one thing, it could hold
both objects to avoid slipping. For the other, it could minimize the necessary
deliberate motion design or motion planning by switching grippers using in-hand
commands.

Second, the hand must be able to align objects along a common axis. Before
inserting a peg into a hole, a process is needed to calibrate the peg and the
hole using vision systems or spiral/linear search. The calibration and searching
process is time consuming. A better solution could be holding the two objects,
align them in-hand to correct the positions of a peg and a hole. Aligning
objects along a common axis is not used in the human demonstration in
Fig.\ref{fig:steps1} and Fig.\ref{fig:steps2}.

Third, a robotic hand that can perform similar tasks must be able to perform
in-hand manipulation like pull-in and push out. Fig. \ref{fig:steps2}(b) shows
that because the retainer pin is short, it is pushed against the idle pulley by
one finger while another two fingers grasp the pulley to prepare for inserting
into the spacer. The process cannot be done by a single gripper or
grippers installed on multiple robotic arms like \cite{dogar2015multi},
\cite{wan2017teaching}. Instead, the gripper needs to be able to perform in-hand
manipulation. It needs to grasp both the retainer pin and the idle pulley, push
the retainer pin into the pulley using in-hand manipulation, and switch to
multi-object holding mode before inserting them into the spacer and slot.

These analysis led us to explore the possibility of performing insertion of a
peg into a pulley using a double jaw design.
Our expectation is two robotic hands, one is the double jaw hand, the other is
a simple gripper, are installed to a dual-arm robot to perform
peg-in-multi-hole tasks. The double jaw hand could hold two objects, align them,
and assemble those two objects into a complex using in-hand pull-in and
push-out. It could also cooperate with the simple gripper on another arm to
perform similar peg-in-hole tasks and insert the complex into other holes held.

\section{Gripper Design}\label{gripper_design}
In this section, an explanation of the design of the double jaw hand is
given. The double jaw hand is consisted of two parallel
grippers that are controlled independently, share the same approaching
vector, and are connected by a prismatic joint.

The hand has four DoFs driven by four motors. Robotis XM430-350-R servo motors
were used because they have a high maximum torque of 4.1Nm (12V), a
relatively small size, and a built-in
control system that allows users switch between position, velocity, and force
control. Details of the inner gripper, the outer gripper, and the
prismatic joint are as follows.

\subsection{Inner gripper}
\begin{figure}[!htbp]
\includegraphics[width=1.0\columnwidth]{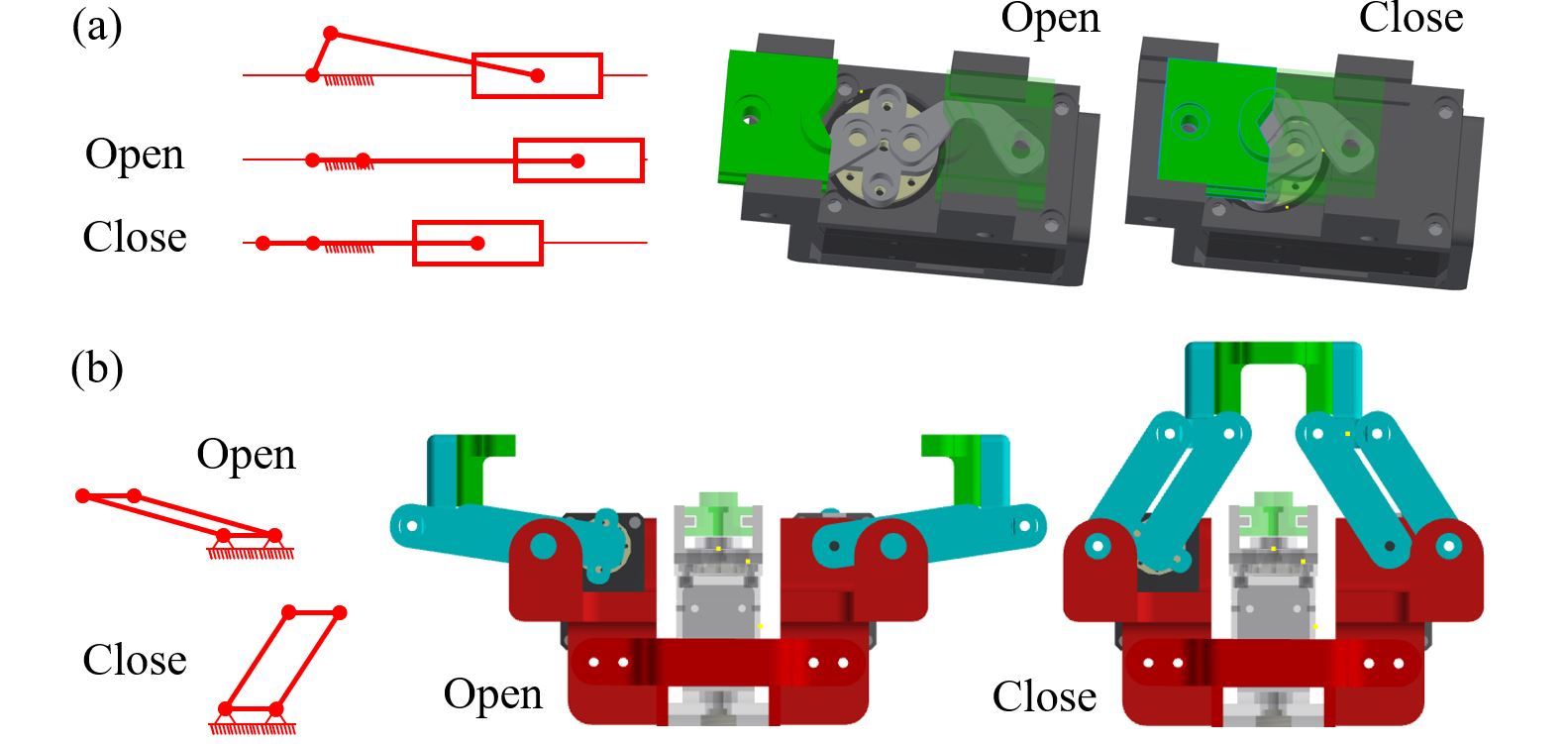}
\caption{(a) Mechanism of the inner gripper. It is essentially a slide crank
shown in the left.
(b) Mechanism of the outer gripper. It is essentially a parallelogram.}
\label{fig:open_close}
\end{figure}

The inner gripper is a parallel gripper controlled using one Robotis XM430-350-R
motor. The opening and closing motion is shown on Fig.
\ref{fig:open_close}(a). The gripper is actuated by a crank-slider mechanism
which is simple and transmits a large force for holding objects firmly.

The maximum stroke of the inner gripper is about 20 mm. The maximum stroke is
small because it is made to precisely and firmly grasp objects like the pulley
shaft and the retainer pin shown in Fig.\ref{fig:objects}. Here, the pulley
shaft has a diameter of 10 mm. The retainer pin has a diameter of 6 mm and
larger head with a diameter of 9 mm. 

The fingertips of the inner gripper are cut into v-shapes since the target
objects are generally cylindrical in shape. The v-shape fingertips also help to
align objects. In addition, to account for the larger head of the retainer pin,
the bottom part of the fingertips has a slightly larger v-shape cut.
    
\subsection{Outer gripper}\label{outergripper}
The outer gripper is also a parallel gripper. The mechanism used for
transmission is a parallelogram. The gripper has two DoFs. Each finger is
controlled using one Robotis XM430-350-R motor, and the two fingers can move
independently. Fig. \ref{fig:open_close}(b) shows the open and close motion of
the outer gripper. Although not shown in the figure, the two fingers can also
move independently. The stroke of the outer gripper is 150 mm, which is able
to hold the large pulleys shown in Fig.\ref{fig:objects}. The fingertips of
the outer gripper are also cut into v-shapes to align cylindrical objects.

There are two main reasons for making the outer gripper has two DoF. First, it
is to keep the middle part empty. The middle part must be left empty to prepare
space for the prismatic joint to connect inner and outer gripper. 
Second, it is to keep the hand symmetric in mass distribution. Symmetric design
will facilitate impedance control in dual-arm assembly. Third,
2DoFs allow non-synchronous motion, which might be useful for later
study. It is therefore advisable to install two independent motors sideways.

\subsection{Prismatic Joint}
\begin{figure}[!htbp]
\includegraphics[width=\columnwidth]{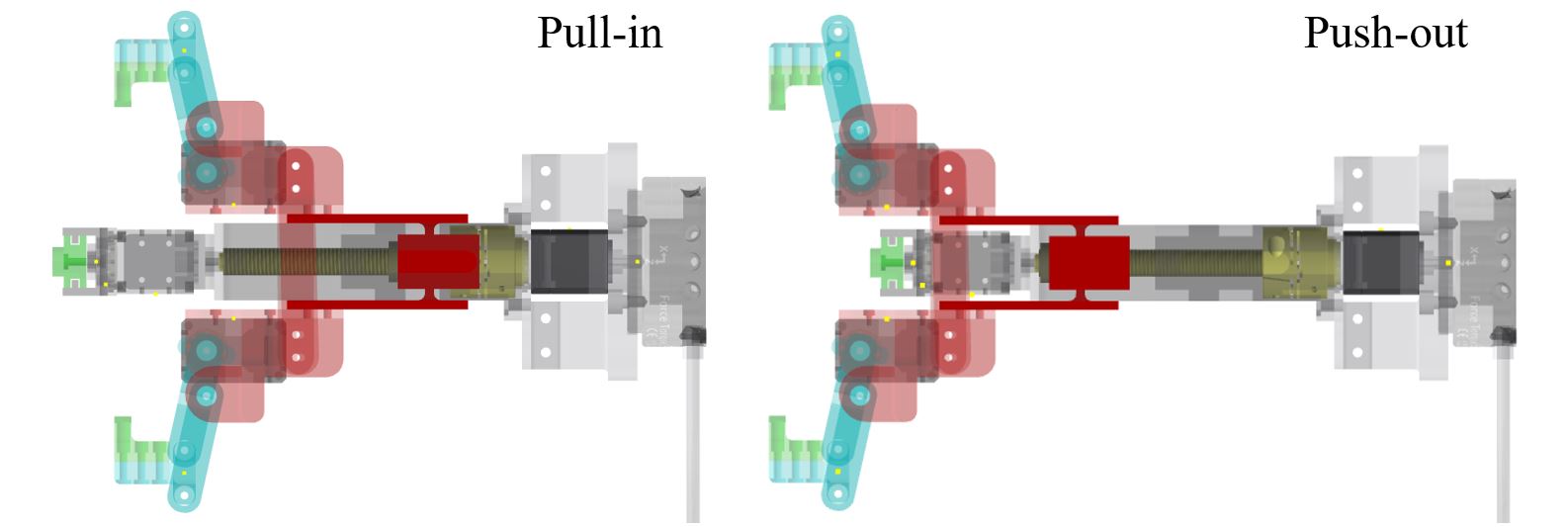}
\caption{Motion of the prismatic joint. The prismatic joint is essentially a
linear screw.}
\label{fig:prismatic}
\end{figure}

The prismatic joint is used to connect the two parallel grippers together along
the same approaching vector. We added a prismatic joint because this allows
for the two grippers to have one DoF relative to each other. The addition of a
prismatic joint enables in-hand pull-in and push-out, making one hand
peg-in-hole possible.

\begin{figure}[!htbp]
\includegraphics[width=\columnwidth]{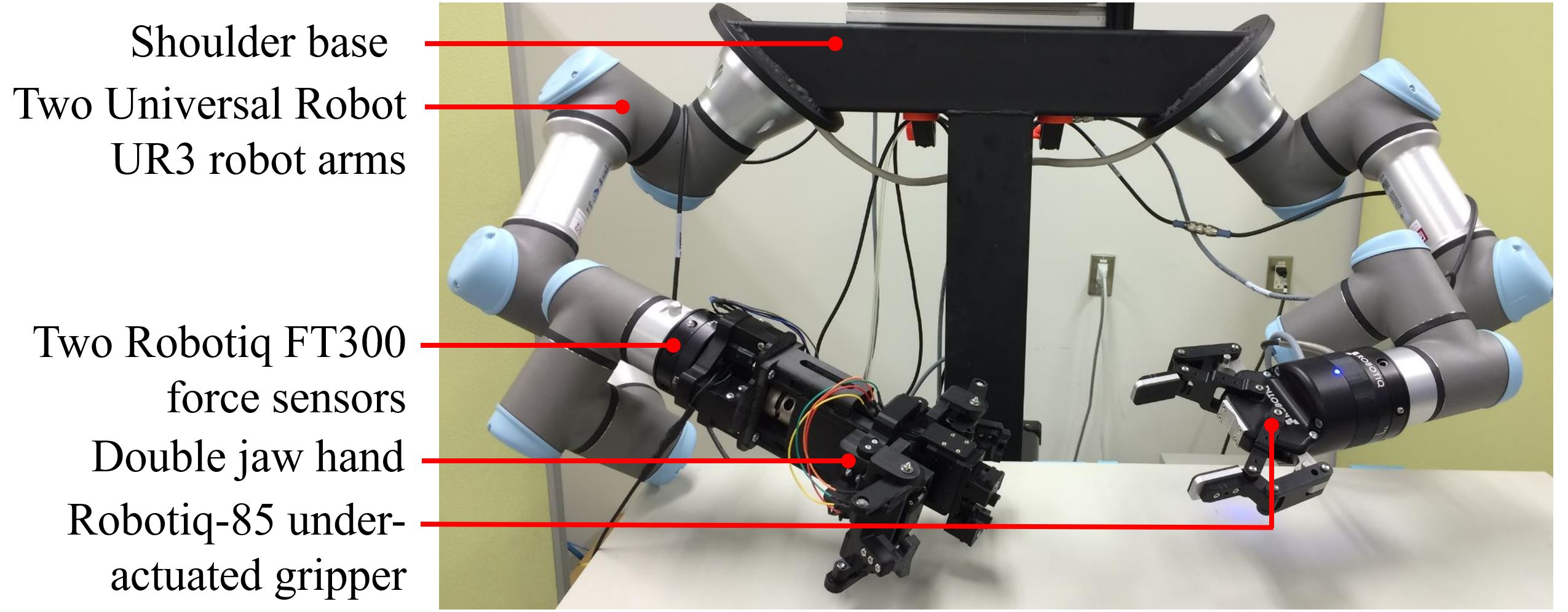}
\caption{Bimanual robot platform used to conduct the experiments. Force sensors
are used for impedance control.}
\label{fig:bimanual_robot}
\end{figure}
\begin{figure*}
	\centering
	\includegraphics[width=.97\textwidth]{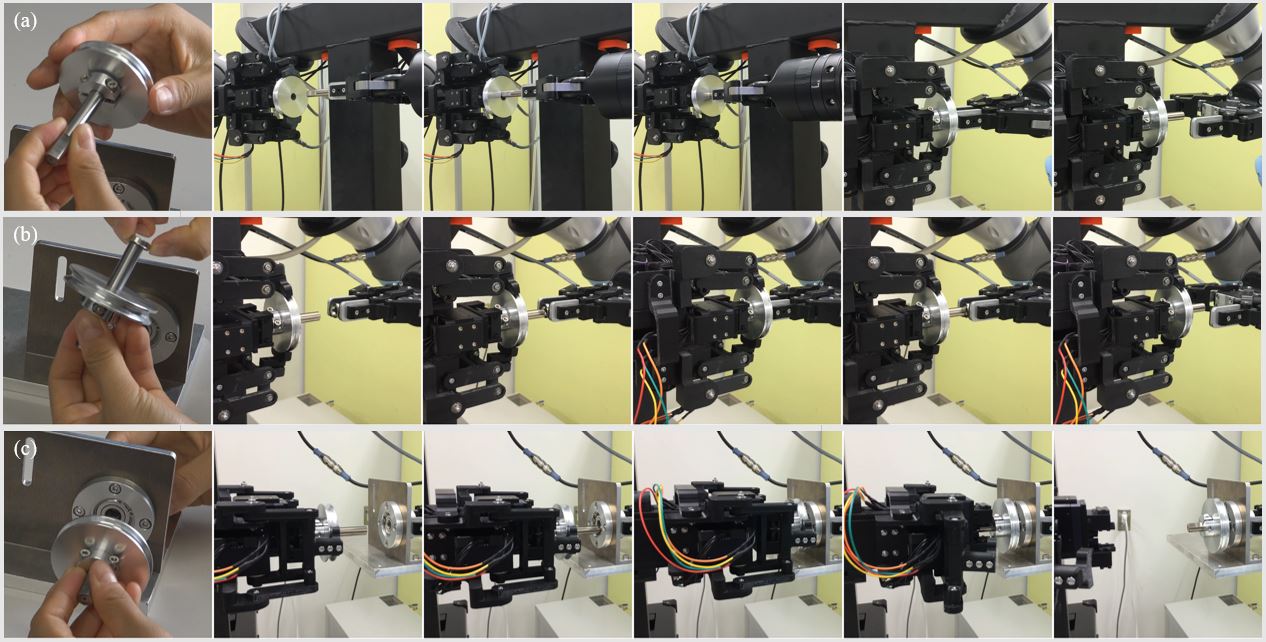}
	\caption{Peg-in-multi-hole assembly experiments on the clamping pulley set.
	(a) Insertion of the pulley shaft into the clamping pulley. (b) Insertion of
	a complex (the pulley shaft and the clamping pulley) into the pulley
	shaft spacer.
	(c) Insertion of a complex (the pulley shaft, the clamping pulley, and
	the pulley shaft spacer) into the bearing on the base.}
	\label{fig:pulleyshaftexperiment}
\end{figure*}

The mechanism of the prismatic joint is a linear screw. The nut of the linear
screw is connected to the outer gripper. The lead is coupled to one
motor. Fig.\ref{fig:prismatic} shows the motion of the prismatic joint.
The Robotis XM430-350-R motor allows close-loop control, which can stop the
linear screw in the presence of mechanical stoppers. The prismatic joint is
calibrated such that 0 mm refers to the position when the outer gripper is
completely pulled-in (left, Fig.\ref{fig:prismatic}). 73 mm refers to the
maximum position when the prismatic joint is completely pushed-out (right,
Fig.\ref{fig:prismatic}).

\section{Experiments and Discussion}\label{experiments}
Two sets of peg-in-multi-hole tasks in the WRC2018 assembly challenge are
performed using a bimanual robot. Fig.\ref{fig:bimanual_robot} shows the
configurations of the bimanual robot. It comprises
two Universal Robot UR3 robotic arms attached to a shoulder base.
Two Robotiq FT300 force sensors were attached to the end of the two arms. The
double jaw hand developed in this paper is attached to the force sensor on
to the right arm. The Robotiq-85 two-finger adaptive gripper is attached to
the force sensor on the left arm.

In this section we present the details of peg-in-multi-hole
experiments performed by the dual arm robot using our dual jaw hand. We also
discuss about the experiments of using the
double jaw hand to insert a peg into a pulley with in-hand manipulation.

\subsection{Experiments}\label{sub_experiment}
\subsubsection{Peg-in-multi-hole assembly of the clamping pulley set and the
idle pulley set}\label{clampingpulleyexperiment} This subsection shows an
explanation on the results of the two sets of peg-in-multi-hole assembly. The
experimental flow is as follows. First, we manually made the double jaw
hand and the parallel gripper grasp the objects for the targeted peg-in-hole
assembly. After that, we ran a routine which consists of linear search, spiral
search, and insertion using impedance control to push the pegs into holes.
Linear search means making the gripper holding a peg moves towards the other
gripper until the peg hits an obstacle. The robot stops the linear motion in
the presence of obstacles.
Spiral search refers to the motion of repeatedly trying to locate a
hole while changing the position of the end of the peg. The robot
draws circles of increasingly larger radius until it finds a hole. This is
necessary to account for imprecision. Once the robot finds the location of the
hole, it inserts the peg into the hole using impedance control. The
insertion process stops until the robot are moved to a given pose.
After successful insertion, we manually controlled the motions of the two
grippers and prepare them for the following complex-in-hole assembly.

It is true that the entire motion sequences, including picking objects from a
given spot accurately with the help of visual sensors, is an important part of
the entire peg-in-multi-hole assembly. Nevertheless, given that the focus of these
experiments were to test the capability of the double jaw hand to grasp
multiple objects and manipulate these objects along a common axis using a
prismatic joint, the picking parts of the experiments are to be performed as
part of the future work. Similarly, while automation of motions of the robot and
the grippers is important part of the research, its automation does not directly
relate to the focus of the paper, therefore we decided to do it in the future.
One aspect that was not manually adjusted was the position of the robot arms
after it moved to the starting position prior to a peg-in-hole motion. This was
to make the spiral search observable.

Fig.\ref{fig:pulleyshaftexperiment} shows
images sequences of the robot performing the peg-in-multi-hole assembly of the
clamping pulley set. In Fig.\ref{fig:pulleyshaftexperiment}(a), the parallel
gripper installed to the left arm holds the pulley shaft and inserts it
to a clamping pulley held by the double jaw hand to minimize the need for two
handovers. In the Fig.\ref{fig:pulleyshaftexperiment}(b, c), the shaft-pulley
complex is held by the double jaw hand. The double jaw hand inserts the complex
into the hole of a pulley shaft spacer and the hole of a bearing.

\begin{figure*}
  \centering
  \includegraphics[width=.97\textwidth]{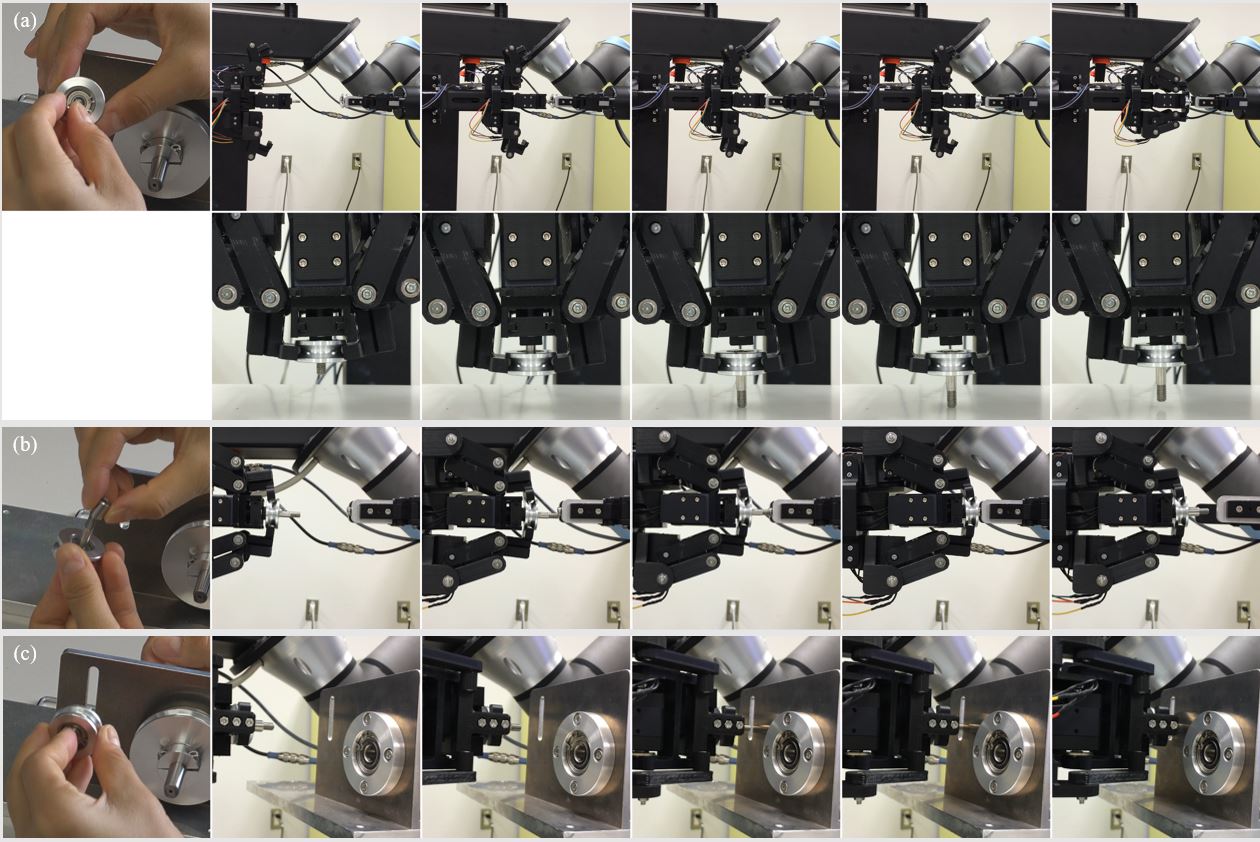}
  \caption{Peg-in-multi-hole assembly experiments on the idle pulley set. (a)
  Upper row: Insertion of the retainer pin into the idle pulley; Lower row:
  In-hand manipulation to prepare for the next peg-in-hole assembly. (b)
  Insertion of a complex (the retainer pin and the idle pulley) into the
  retainer pin spacer. (c) Insertion of a complex (the retainer pin, the
  idle pulley, the retainer pin spacer) into a slot on the base.}
  \label{fig:retainerpinexperiment}
\end{figure*}

Fig.\ref{fig:retainerpinexperiment} shows image sequences of the robot
performing peg-in-multi-hole assembly of the idle pulley set.
Compared to inserting the pulley shaft into the clamping pulley, inserting the
retainer pin to the idle pulley was more complicated since the pin must be fully
embedded in the pulley. The process is divided into two steps, including an
insertion step and an in-hand manipulation step (from grasping the retainer pin
to pushing the retainer pin). Fig.\ref{fig:retainerpinexperiment}(a) shows the
two steps. 
The two complex-in-hole assembly in Fig.\ref{fig:retainerpinexperiment}(b, c)
are similar as the ones from the clamping pulley set.

\begin{figure*}
  \includegraphics[width=.97\textwidth]{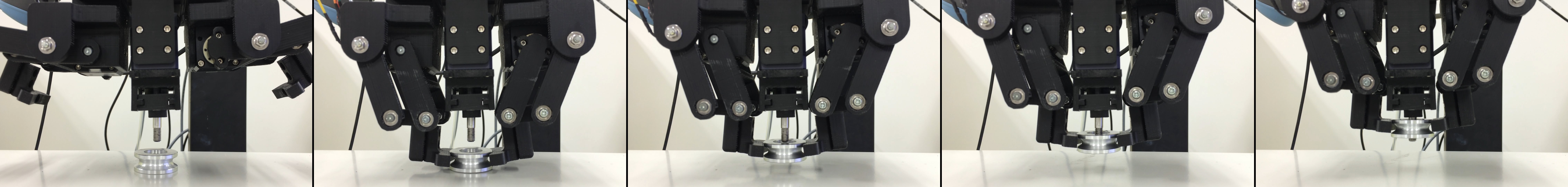}
  \centering
  \caption{Insertion of the retainer pin to the idle pulley using in-hand
  manipulation.}
  \label{fig:onehand}
\end{figure*}

The results of the experiments after performing multiple trials of the
peg-in-multi-hole assembly are as follows. The first step of the three-step
peg-in-multi-hole assembly is generally successful, even though it takes
some time for the robot to locate a hole. The second
step of the peg-in-multi-hole assembly is the one which may fail depending on
how the spacer becomes increasingly tilted every time the peg hits it during
spiral search. The third step is also
generally successful.

\subsubsection{Insertion of a peg to a pulley using the double jaw
gripper}\label{onehandexperiment}
In this subsection, we examine the possibility of using double jaw hand
to align a peg and a pulley and perform insertion by pulling-in prismatic joint.
Fig.\ref{fig:onehand} shows the result of the experiments performed on the
retainer pin and the idle pulley. We managed to conduct this experiment
successfully repeatedly with the retainer pin and the idle pulley, but the
experiments are largely not successful for the pulley shaft and the clamping
pulley. This is probably because the clamp part and the pulley part of the
clamping pulley have different hole sizes. The insertion is performed such that the peg
entered first to the protruding clamp part, otherwise it would end up requiring
handovers. While insertion into the clamp part is largely successful, the peg
easily hits the pulley part and gets stuck, leading to failure in assembly. This
result is like our expectation because it shares the similar reason as to why
the insertion such as in Fig \ref{fig:pulleyshaftexperiment}(a) is chosen
instead of insertion to the other side (clamp part) of the clamping pulley.





\subsection{Discussion} \label{discussion}
First, we discuss how the experiments verified the
requirements on gripper design stated in Section \ref{introduction} and Section
\ref{problem_analysis}. Throughout the experiments, the hand is performing
tasks like complex-in-hole, demonstrating that the hand could hold two objects.
At the end of the first peg-in-hole between the retainer pin and the idle pulley, the double jaw hand held both objects. The gripper continued to
release the retainer pin and push the retainer pin to keep it in place while
holding the idle pulley. This shows the intrinsic
in-hand manipulation capability of the gripper. The clearance between the
retainer pin and the idler pulley is 0.01 mm, indicating that the hand is good
at aligning objects. The outer gripper also has enough stroke to hold large
objects.

Second, we discuss the shortcomings of the developed gripper. The two grippers
of the double jaw hand are connected by a linear
screw. Given that predefined settings in Robotis XM430-350-R do not allow users
to control the rotational speed in the current-based position control mode which
is used to control the prismatic joint, the prismatic joint movement is slow. It
takes about 55 seconds to move from 0 mm to 73 mm, which means that the linear
speed is approximately 1.3 mm/second. This restricts the efficiency of
peg-in-hole assembly using in-hand pull-in and push-out.

\section{Conclusions and Future Work}\label{conclusion}

This paper presented a novel double jaw hand capable of holding two objects and
performing in-hand manipulation. Experimental and real-world execution results
showed that the hand is simple, robust, and is able to perform peg-in-multi-hole
assembly with the help of an external gripper. Our future work includes optimizing
the size of the double jaw hand and developing automatic grasp planning and
control algorithms.
 



\section*{Acknowledgement}

This paper is based on results obtained from a project commisioned by the New
Energy and Industrial Technology Development Organisation (NEDO).
The authors
would like to thank Felix von Drigalski and Chisato Nakashima from OMRON SINIC
X and fellow members of Robotics Manipulation Laboratory, Osaka University for
their support in the development of this work.


\bibliographystyle{IEEEtran}
\balance
\bibliography{paperJoshua}

\begin{thebibliography}{10}
\providecommand{\url}[1]{#1}
\csname url@samestyle\endcsname
\providecommand{\newblock}{\relax}
\providecommand{\bibinfo}[2]{#2}
\providecommand{\BIBentrySTDinterwordspacing}{\spaceskip=0pt\relax}
\providecommand{\BIBentryALTinterwordstretchfactor}{4}
\providecommand{\BIBentryALTinterwordspacing}{\spaceskip=\fontdimen2\font plus
\BIBentryALTinterwordstretchfactor\fontdimen3\font minus
  \fontdimen4\font\relax}
\providecommand{\BIBforeignlanguage}[2]{{%
\expandafter\ifx\csname l@#1\endcsname\relax
\typeout{** WARNING: IEEEtran.bst: No hyphenation pattern has been}%
\typeout{** loaded for the language `#1'. Using the pattern for}%
\typeout{** the default language instead.}%
\else
\language=\csname l@#1\endcsname
\fi
#2}}
\providecommand{\BIBdecl}{\relax}
\BIBdecl

\bibitem{hormann1991development}
A.~Hormann and U.~Rembold, ``Development of an advanced robot for autonomous
  assembly,'' in \emph{Robotics and Automation, 1991. Proceedings., 1991 IEEE
  International Conference on}.\hskip 1em plus 0.5em minus 0.4em\relax IEEE,
  1991, pp. 2452--2457.

\bibitem{dogar2015multi}
M.~Dogar, A.~Spielberg, S.~Baker, and D.~Rus, ``Multi-robot grasp planning for
  sequential assembly operations,'' in \emph{Robotics and Automation (ICRA),
  2015 IEEE International Conference on}.\hskip 1em plus 0.5em minus
  0.4em\relax IEEE, 2015, pp. 193--200.

\bibitem{wan2017teaching}
W.~Wan, F.~Lu, Z.~Wu, and K.~Harada, ``Teaching robots to do object assembly
  using multi-modal 3d vision,'' \emph{Neurocomputing}, vol. 259, pp. 85--93,
  2017.

\bibitem{monkman2007robot}
G.~J. Monkman, S.~Hesse, R.~Steinmann, and H.~Schunk, \emph{Robot
  grippers}.\hskip 1em plus 0.5em minus 0.4em\relax John Wiley \& Sons, 2007.

\bibitem{Mason01}
M.~T. Mason, \emph{{Mechanics of Robotic Manipulation}}.\hskip 1em plus 0.5em
  minus 0.4em\relax The MIT Press, 2001.

\bibitem{zeng2017robotic}
A.~Zeng, S.~Song, K.-T. Yu, E.~Donlon, F.~R. Hogan, M.~Bauza, D.~Ma, O.~Taylor,
  M.~Liu, E.~Romo \emph{et~al.}, ``Robotic pick-and-place of novel objects in
  clutter with multi-affordance grasping and cross-domain image matching,''
  \emph{arXiv preprint arXiv:1710.01330}, 2017.

\bibitem{atakuru2018robotic}
T.~Atakuru and E.~Samur, ``A robotic gripper for picking up two objects
  simultaneously,'' \emph{Mechanism and Machine Theory}, vol. 121, pp.
  583--597, 2018.

\bibitem{xu2016design}
Z.~Xu and E.~Todorov, ``Design of a highly biomimetic anthropomorphic robotic
  hand towards artificial limb regeneration,'' in \emph{Robotics and Automation
  (ICRA), 2016 IEEE International Conference on}.\hskip 1em plus 0.5em minus
  0.4em\relax IEEE, 2016, pp. 3485--3492.

\bibitem{townsend2000barretthand}
W.~Townsend, ``The barretthand grasper--programmably flexible part handling and
  assembly,'' \emph{Industrial Robot: an international journal}, vol.~27,
  no.~3, pp. 181--188, 2000.

\bibitem{massa2002design}
B.~Massa, S.~Roccella, M.~C. Carrozza, and P.~Dario, ``Design and development
  of an underactuated prosthetic hand,'' in \emph{Robotics and Automation,
  2002. Proceedings. ICRA'02. IEEE International Conference on}, vol.~4.\hskip
  1em plus 0.5em minus 0.4em\relax IEEE, 2002, pp. 3374--3379.

\bibitem{deimel2016novel}
R.~Deimel and O.~Brock, ``A novel type of compliant and underactuated robotic
  hand for dexterous grasping,'' \emph{The International Journal of Robotics
  Research}, vol.~35, no. 1-3, pp. 161--185, 2016.

\bibitem{cannella2013design}
F.~Cannella, F.~Chen, C.~Canali, A.~Eytan, A.~Bottero, and D.~Caldwell,
  ``Design of an industrial robotic gripper for precise twisting and
  positioning in high-speed assembly,'' in \emph{System Integration (SII), 2013
  IEEE/SICE International Symposium on}.\hskip 1em plus 0.5em minus 0.4em\relax
  IEEE, 2013, pp. 443--448.

\bibitem{chen2014hand}
F.~Chen, F.~Cannella, C.~Canali, T.~Hauptman, G.~Sofia, and D.~Caldwell,
  ``In-hand precise twisting and positioning by a novel dexterous robotic
  gripper for industrial high-speed assembly,'' in \emph{Robotics and
  Automation (ICRA), 2014 IEEE International Conference on}.\hskip 1em plus
  0.5em minus 0.4em\relax IEEE, 2014, pp. 270--275.

\bibitem{ma2016m}
R.~R. Ma, A.~Spiers, and A.~M. Dollar, ``M 2 gripper: Extending the dexterity
  of a simple, underactuated gripper,'' in \emph{Advances in reconfigurable
  mechanisms and robots II}.\hskip 1em plus 0.5em minus 0.4em\relax Springer,
  2016, pp. 795--805.

\bibitem{odhner2014compliant}
L.~U. Odhner, L.~P. Jentoft, M.~R. Claffee, N.~Corson, Y.~Tenzer, R.~R. Ma,
  M.~Buehler, R.~Kohout, R.~D. Howe, and A.~M. Dollar, ``A compliant,
  underactuated hand for robust manipulation,'' \emph{The International Journal
  of Robotics Research}, vol.~33, no.~5, pp. 736--752, 2014.

\bibitem{yamaguchi2013development}
K.~Yamaguchi, Y.~Hirata, and K.~Kosuge, ``Development of robot hand with
  suction mechanism for robust and dexterous grasping,'' in \emph{Intelligent
  Robots and Systems (IROS), 2013 IEEE/RSJ International Conference on}.\hskip
  1em plus 0.5em minus 0.4em\relax IEEE, 2013, pp. 5500--5505.

\bibitem{kakogawa2016underactuated}
A.~Kakogawa, H.~Nishimura, and S.~Ma, ``Underactuated modular finger with
  pull-in mechanism for a robotic gripper,'' in \emph{Robotics and Biomimetics
  (ROBIO), 2016 IEEE International Conference on}.\hskip 1em plus 0.5em minus
  0.4em\relax IEEE, 2016, pp. 556--561.

\bibitem{dafle2014extrinsic}
N.~Chavan-Dafle, A.~Rodriguez, R.~Paolini, B.~Tang, S.~S. Srinivasa,
  M.~Erdmann, M.~T. Mason, I.~Lundberg, H.~Staab, and T.~Fuhlbrigge,
  ``Extrinsic dexterity: In-hand manipulation with external forces,'' in
  \emph{Robotics and Automation (ICRA), 2014 IEEE International Conference
  on}.\hskip 1em plus 0.5em minus 0.4em\relax IEEE, 2014, pp. 1578--1585.

\bibitem{rodriguez2013effector}
A.~Rodriguez and M.~T. Mason, ``Effector form design for 1dof planar
  actuation,'' in \emph{Robotics and Automation (ICRA), 2013 IEEE International
  Conference on}.\hskip 1em plus 0.5em minus 0.4em\relax IEEE, 2013, pp.
  349--356.

\bibitem{inoue1971computer}
H.~Inoue, ``Computer controlled bilateral manipulator,'' \emph{Bulletin of
  JSME}, vol.~14, no.~69, pp. 199--207, 1971.

\bibitem{mason1981compliance}
M.~T. Mason, ``Compliance and force control for computer controlled
  manipulators,'' \emph{IEEE Transactions on Systems, Man, and Cybernetics},
  vol.~11, no.~6, pp. 418--432, 1981.

\bibitem{zheng2017peg}
Y.~Zheng, X.~Zhang, Y.~Chen, and Y.~Huang, ``Peg-in-hole assembly based on
  hybrid vision/force guidance and dual-arm coordination,'' in \emph{Robotics
  and Biomimetics (ROBIO), 2017 IEEE International Conference on}.\hskip 1em
  plus 0.5em minus 0.4em\relax IEEE, 2017, pp. 418--423.

\end{thebibliography}

\end{document}